\documentclass[sigconf, review=false]{acmart}

\usepackage{booktabs} 

\setcopyright{acmcopyright}

\copyrightyear{2022}
\acmYear{2022}
\setcopyright{othergov}\acmConference[ICVGIP'22]{Proceedings of the Thirteenth Indian Conference on Computer Vision, Graphics and Image Processing}{December 8--10, 2022}{Gandhinagar, India}
\acmBooktitle{Proceedings of the Thirteenth Indian Conference on Computer Vision, Graphics and Image Processing (ICVGIP'22), December 8--10, 2022, Gandhinagar, India}
\acmPrice{15.00}
\acmDOI{10.1145/3571600.3571607}
\acmISBN{978-1-4503-9822-0/22/12}
\begin{document}
\title{Convolutional Ensembling based Few-Shot Defect Detection Technique}

\author{Soumyajit Karmakar}
\affiliation{%
  \institution{Indian Institute of Information Technology Guwahati (IIITG)}
  \city{Guwahati}
  \state{Assam}
  \country{India}
  \postcode{781015}
}

\author{Abeer Banerjee}
\affiliation{%
  \institution{CSIR-Central Electronics Engineering Research Institute (CSIR-CEERI)}
  \city{Pilani}
  \state{Rajasthan}
  \country{India}
  \postcode{333031}
}

\author{Prashant Sadashiv Gidde}
\affiliation{%
  \institution{CSIR-Central Electronics Engineering Research Institute (CSIR-CEERI)}
  \city{Pilani}
  \state{Rajasthan}
  \country{India}
  \postcode{333031}
}

\author{Sumeet Saurav}
\affiliation{%
  \institution{CSIR-Central Electronics Engineering Research Institute (CSIR-CEERI)}
  \city{Pilani}
  \state{Rajasthan}
  \country{India}
  \postcode{333031}
}

\author{Sanjay Singh}
\affiliation{%
  \institution{CSIR-Central Electronics Engineering Research Institute (CSIR-CEERI)}
  \city{Pilani}
  \state{Rajasthan}
  \country{India}
  \postcode{333031}
}

\renewcommand{\shortauthors}{}

\begin{abstract}
Over the past few years, there has been a significant improvement in the domain of few-shot learning. This learning paradigm has shown promising results for the challenging problem of anomaly detection, where the general task is to deal with heavy class imbalance. Our paper presents a new approach to few-shot classification, where we employ the knowledge base of multiple pre-trained convolutional models that act as the backbone for our proposed few-shot framework. Our framework uses a novel ensembling technique for boosting the accuracy while drastically decreasing the total parameter count, thus paving the way for real-time implementation. We perform an extensive hyperparameter search using a power-line defect detection dataset and obtain an accuracy of 92.30\% for the $5$-way $5$-shot task. Without further tuning, we evaluate our model on competing standards with the existing state-of-the-art methods and outperform them. 
\end{abstract}

%
%



\maketitle

\pagestyle{plain}

\section{Introduction}
In conventional deep-learning-based computer vision approaches, one can observe a positive relationship between the size of the training dataset and the performance of the model. In contrast, the few-shot based approaches attempt to achieve similar performances while using significantly lesser training dataset.

There has been a lot of recent research in this domain \cite{fei2006one, lee2019meta, li2019finding, finn2017model, finn2018probabilistic, yoon2018bayesian, nichol2018reptile, lee2018gradient}. The major benefit of a few-shot learning based solution to a computer vision problem, say image classification, is that the overall computation cost of achieving a certain level of accuracy is drastically lower as compared to traditional data-driven approaches. As few-shot based approach requires a few examples of a per class by definition, the task of data collection and annotation becomes significantly easier. This attribute makes it perfectly suitable for dealing with problems where the collection of data for a particular class is either difficult or the event in consideration is naturally rare. Researchers have been employing few-shot learning in the problem of anomaly detection \cite{lu2020few, ding2021few, sheynin2021hierarchical} as anomalies are naturally rare, which creates a huge class imbalance. While there have been approaches to solve the problem of class imbalance with synthetic data generation using generative adversarial networks \cite{qasim2020red, yang2021ida}, they still suffer from drawbacks such as, huge computational cost involved during training.    

One of the most popular frameworks for this task is meta-learning. Here the model focuses on learning to learn, rather than memorizing the particular features of images. This helps in enabling the model to distinguish objects without the requirement of a huge dataset. A few-shot problem is usually defined by the pair $n$-way $k$-shot, where $n$ refers to the number of classes in question and $k$ refers to the number of examples in each class on which the model is trained. The training set thus formed is called the support set, and the testing set is called the query set.

For training the few-shot learner, two commonly used approaches are gradient-based approach and metric-based approach. In the case of gradient-based approaches, the base model is updated as a trainable function \cite{bengio1992global} and the gradients are then back-propagated across it \cite{maclaurin2015gradient, finn2017model}. In metric-based approaches, a feature embedding is learnt which is then used to classify the query images based on a similarity function \cite{gidaris2018dynamic, sung2018learning}.

\begin{figure*}[t]
    \centering
    \includegraphics[width=1.0\textwidth]{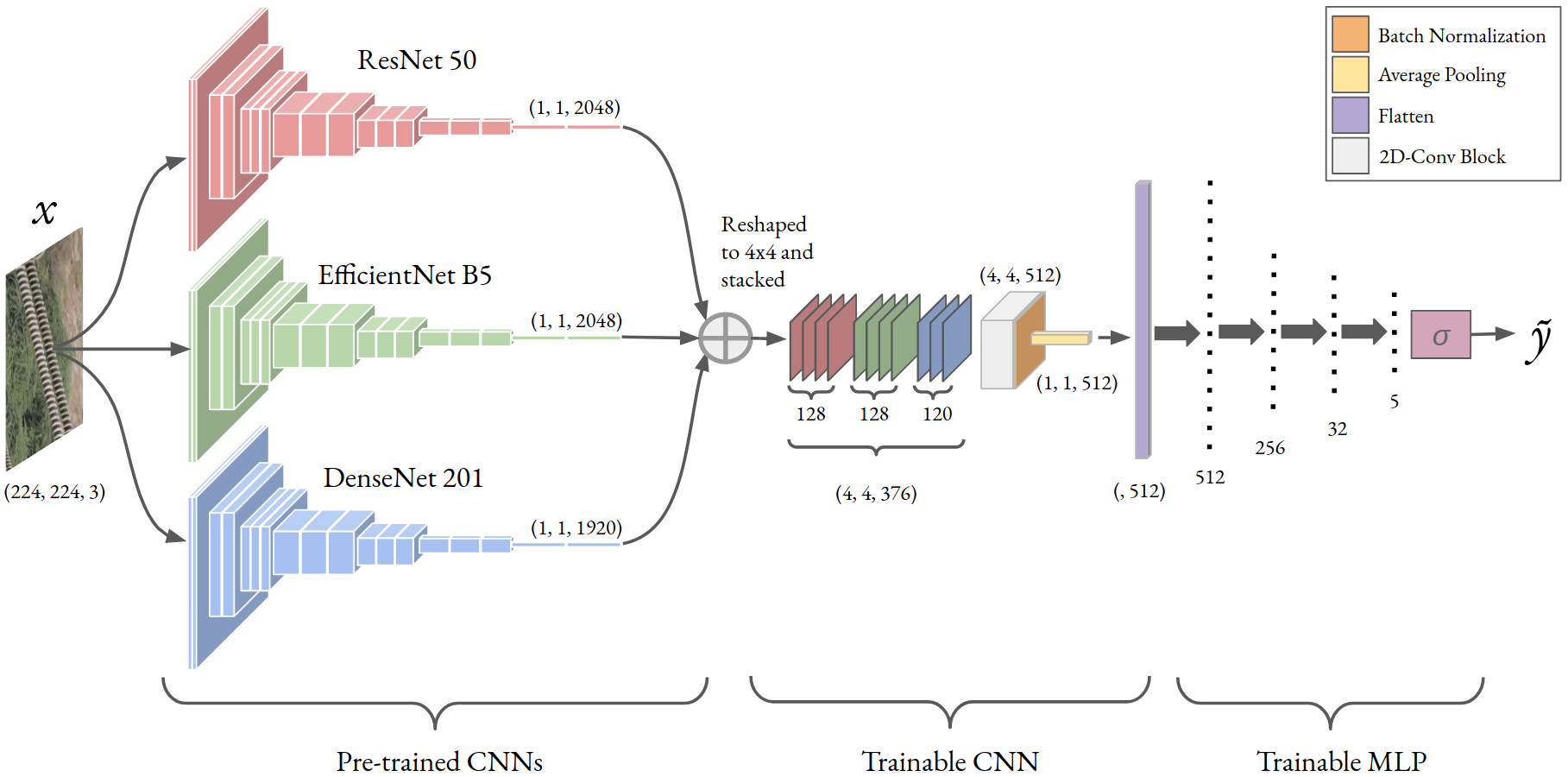} 
    \caption{Proposed Architecture: Three pre-trained backbone feature extractors compute the features of an input image which are ensembled and fed to the trainable CNN-MLP network for few-shot classification.}
    \label{arch}
\end{figure*}

In this paper, we present a new approach, influenced by the work of Chowdhury et al. \cite{chowdhury2021few}. Chowdhury et al. employed a huge combination of library-learners which are basically pre-trained CNNs available off-the-shelf, and used them to parallelly compute the feature embeddings. They combined the features using simple ensembling techniques and passed it to a multi-layered perceptron. Their approach simplifies the training process but involves a huge number of parameters to achieve an acceptable amount of accuracy. We develop a new ensembling strategy that involves the use of a convolutional block to stack and combine the features obtained from each feature extractor. This drastically reduces the parameter counts and boosts the classification accuracy.

We evaluate the reliability of our model on some of the standard datasets namely, the Aircraft \cite{maji2013fine}, Traffic \cite{oreshkin2018tadam}, Omniglot \cite{lake2015human}, FC100 \cite{houben2013detection}, VGG Flower \cite{nilsback2008automated}, Texture \cite{cimpoi2014describing}, and the Powerline Components dataset. This dataset consists of five classes, i.e., $n = 5$, namely the spacer, bolts, nests, insulators, and the defect class, missing bolts. We captured the dataset using a camera-drone flying over industrial power line structures and henceforth in this paper, we will be referring to it as the powerline dataset. This dataset has huge class imbalance, it contains thousands of images of varying resolutions in all classes except in the anomaly class, missing bolts, which contains significantly less number of samples. This dataset emulates the practical scenarios perfectly as the captured images are mostly of low resolutions, which makes the classification task particularly challenging.

\section{Strategy}
We took some well-known off-the-shelf convolutional neural networks, ResNet \cite{he2016deep}, DenseNet \cite{huang2017densely}, Inception \cite{szegedy2015going}, Xception \cite{chollet2017xception}, EfficientNet \cite{tan2019efficientnet}, all trained on ILSVRC2012 \cite{russakovsky2015imagenet} and discarded the fully-connected layers to obtain their respective convolutional segments. These convolution subnetworks are used to extract and form the feature embeddings corresponding to each image. We reshaped the obtained features to a stack of channels and passed it to our proposed model for the few-shot classification task. We experimented with multiple pre-trained CNNs and found that a combination of three such networks had provided the best results. A detailed performance evaluation report using various off-the-shelf pre-trained models has been provided in Section \ref{abla}. 

The support and query sets were generated by randomly sampling our dataset. The few-shot training was performed using the support set that comprised of a few examples for each class while the few-shot query set was used to evaluate the model performance. The model architecture and the training details are discussed in the subsequent sections.

\begin{table}[t]
\centering
\begin{tabular}{lll}
\hline
\textbf{Layers} & \textbf{Output Shape} &\textbf{Parameters}\\
\hline
ResNet 50   & $1 \times 1 \times 2048$ &  23.6M (Frozen)\\
EfficientNet B5 & $1 \times 1 \times 2048$ & 28.5M (Frozen)\\
DenseNet 201   & $1 \times 1 \times 1920$ &  18.3M (Frozen)\\
\hline
Concat   & $1 \times 1 \times 6016$ &  -\\
Reshape stack   & $4 \times 4 \times 376$ &  -\\
\hline
Conv2D   & $4 \times 4 \times 512$ &  1.7M\\
BatchNorm & $4 \times 4 \times 512$ &  2k\\
AvgPool   & $1 \times 1 \times 512$ &  -\\
Flatten   & $512$ &  -\\
Dense In   & $512$ &  262k\\
Hidden Dense 1  & $256$ &  131k\\
Hidden Dense 2   & $32$ &  8k\\
Dense Out   & $5$ &  0.1k\\
\hline
Total Trainable Parameters & & 2.1M\\
\hline
\end{tabular}
\caption{Network architecture parameters obtained after an extensive hyperparameter search.}
\label{arch_table}
\end{table}
\raggedbottom

\begin{figure*}[t]
    \centering
    \includegraphics[width=1.0\textwidth]{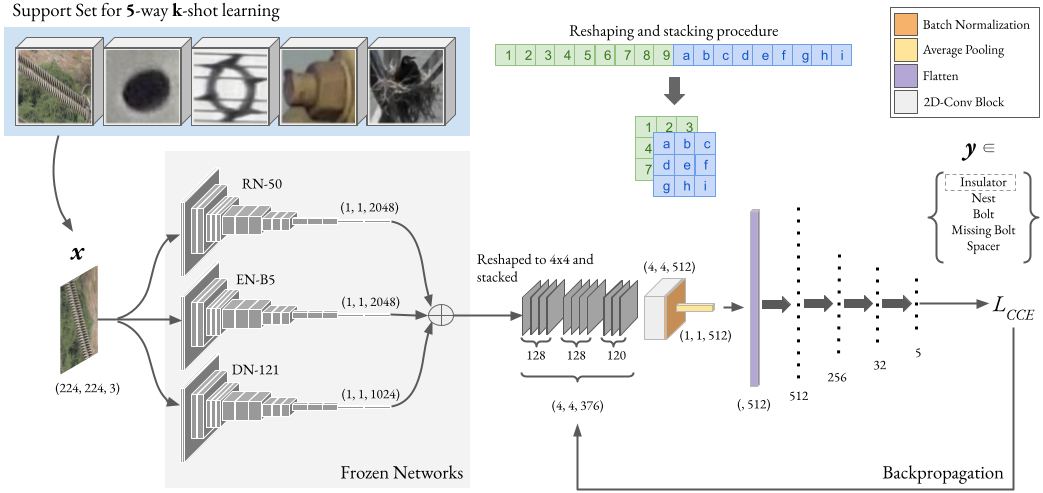} 
    \caption{Training Pipeline: A $5$-way $k$-shot support set is used for the few shot training of the trainable CNN-MLP network. Feature ensembling involves reshaping and stacking of the computed features. Finally, categorical cross-entropy loss, $L_{CCE}$, is backpropagated through the CNN-MLP network.}
    \label{train}
\end{figure*}

\subsection{Architecture}
For our best performing approach, we have used the combination of ResNet 50, EfficientNet B5, and DenseNet 201 for computing the feature embeddings corresponding to each image. The obtained features were reshaped to $4 \times 4$ spatial blocks and stacked to form the input for our proposed model. The stacked channels form the input $x$, that is passed through a convolutional block comprising of [$Conv2D \longrightarrow ReLU \longrightarrow BatchNorm \longrightarrow AvgPool$]. The output of this block is then flattened and passed through a Multi-Layer Perceptron with two hidden layers with 256 neurons and 32 neurons. The output layer has five neurons for providing a 5 class classification which is then passed through a softmax layer to obtain the final classified label $\Tilde{y}$. The detailed architecture is illustrated in Figure \ref{arch} and the model summary  is provided in Table \ref{arch_table}.

\subsection{Training and Implementation}
We used a custom dataset for the few-shot training and evaluation process. We experimented with different numbers of examples per class ($k$) that can be used for training the few shot learner and found that the model has an optimum performance at $k$=5. We randomly selected 32 images from each class from the whole dataset and split them into two groups of 5 and 27 images for the support and query sets, respectively. The support set for the few-shot training process was created by combining the extracted features from the pre-trained networks with their associated labels. The same procedure was followed for the rest of the 27 images, except that their corresponding labels were not supplied, and then the resulting features formed the query set.

Figure \ref{train} illustrates the training pipeline using three pre-trained networks as the feature extractors. We use the proposed feature ensembling strategy to combine the extracted features which are passed to the trainable CNN layers and further propagated to the trainable MLP layer to obtain the classified output. The reshaping and stacking technique is explained visually in the same figure. The network minimizes the categorical cross-entropy loss $L_{CCE}$ which is backpropagated through the trainable layers. We use Adam optimizer \cite{kingma2014adam} with a learning rate of $5\times 10^{-5}$. We use values for the $L_2$ regularization constant as high as $0.5$ to ensure that the model does not overfit to the training data. The network takes 300 epochs for the loss value to saturate, but as the execution time of each epoch is less than a few milliseconds, the overall process does not take more than a few seconds to complete. All training and testing were performed on a system powered by an Intel Xeon 2.90 GHz quad-core CPU with NVIDIA 1080 GPU having 8GB of graphics memory.

\begin{figure*}[t]
    \centering
    \includegraphics[width=1.0\textwidth]{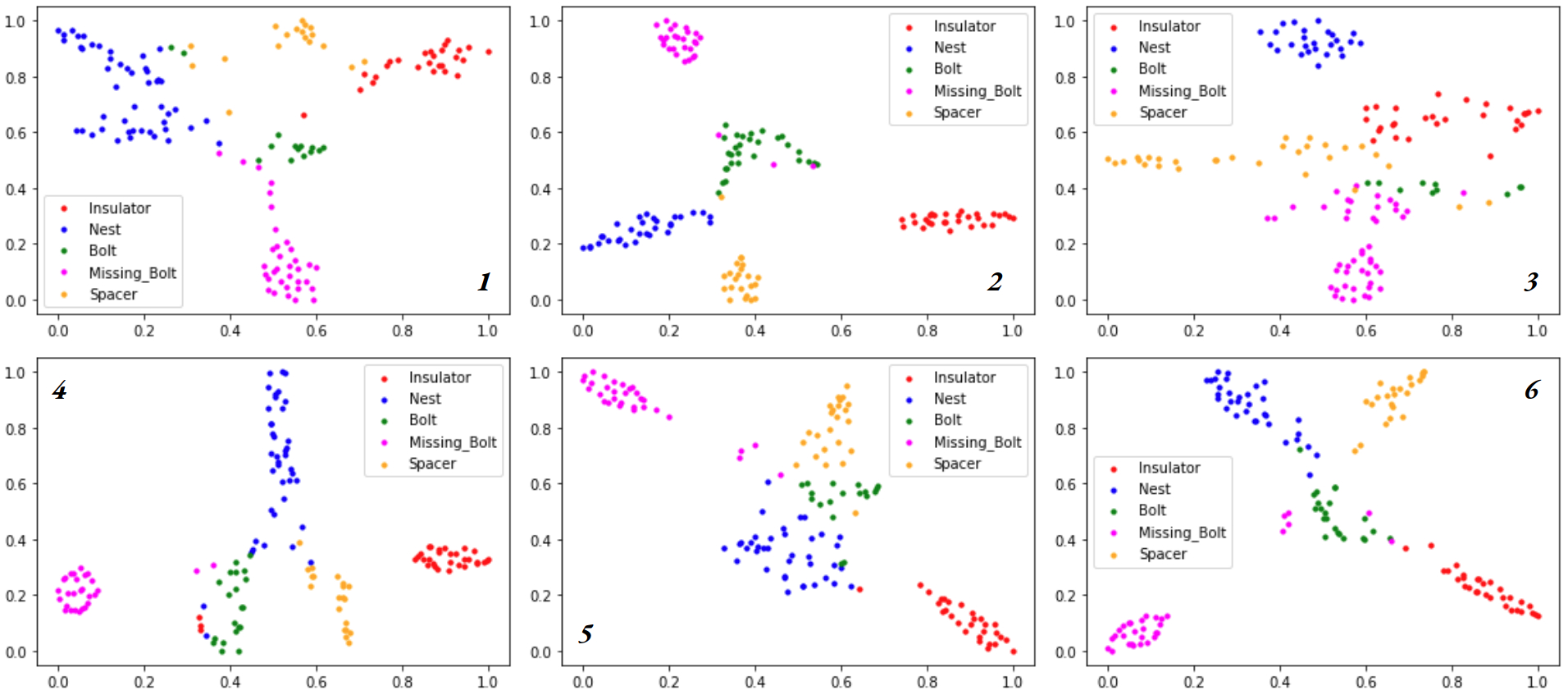} 
    \caption{Ablation study of the proposed framework using TSNE plots: (1) DenseNet 201 only. (2) ResNet 50 only. (3) EfficientNet B5 only. (4) ResNet 50 + DenseNet 201. (5) DenseNet 201 + EfficientNet B5. (6) ResNet 50 + EfficientNet B5.}
    \label{other}
\end{figure*}

\begin{figure}[t]
    \centering
    \includegraphics[width=0.48\textwidth]{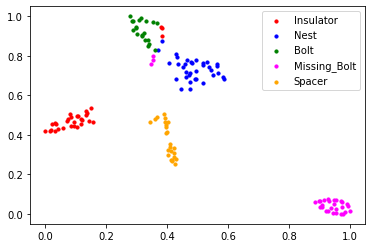} 
    \caption{Ablation study of the proposed framework using TSNE plots: DenseNet 201 + ResNet 50 + EfficientNet B5 combined.}
    \label{full}
\end{figure}

\section{Ablation Study and Hyperparameter Search}\label{abla}

We performed an extensive ablation study to ensure the reliability of our approach. We ablate our model in terms of the feature extractors, the ensembling strategy and the hyperparameters for the trainable CNN and MLP network. We considered eight different backbone networks, each with less than 30M parameters, and 4 different kernel sizes for the ensembling. For studying the ablation of ensembling techniques, we fix the structure of the trainable CNN-MLP model on a trial-error basis to observe the general trend in accuracy. We later on refine this structure based on further ablation studies. For all the testing purposes we have used $k$-fold cross verification to obtain a reliable performance score.

\subsection{Ablation study of ensembling techniques}
We begin with using only one pre-trained network as the backbone. The performance of each of the models corresponding to various kernel sizes for ensembling is presented in Table \ref{abla_ensem}. A clear trend can be observed that the $4\times4$ kernel size works the best for each model. We pick three of the best performing models for further ablation studies.
Since the output shape of most of the networks is either 1024 or 2048, we can easily convert it into 1 or 2 stacks of $32 \times 32$. The output size of DenseNet 201 is 1920 which can only be converted into stacks of $8\times8$ or smaller, therefore, some of the columns in Table \ref{abla_ensem} are missing. For similar reasons, some accuracies corresponding to EfficientNet V2S is also missing. 

\begin{table}[h]
\centering
\begin{tabular}{l l l l l}
\hline
\textbf{Backbone} & \textbf{$32 \times 32$} & \textbf{$16 \times 16$} & \textbf{$8 \times 8$} & \textbf{$4 \times 4$}\\
\hline
ResNet 50 & 87.86 & 86.57 & 85.53 & 88.02\\
ResNet 50 V2 & 82.32 & 85.74 & 86.28 & 87.98\\
DenseNet 121 & 76.39 & 78.47 & 77.78 & 84.22 \\
DenseNet 201 & - & - & 79.04 & 87.53\\
Inception V3 & 64.67 & 63.87 & 69.02 & 75.37\\
Xception & 75.48 & 71.73 & 73.93 & 78.27\\
EfficientNet V2S & - & 78.83 & 78.56 & 82.79\\
EfficientNet B5 & 77.56 & 79.77 & 75.91 & 82.76\\
\hline
\end{tabular}
\caption{Ablation Study of reshaping kernel size for ensembling.}
\label{abla_ensem}
\end{table}
\raggedbottom

\subsection{Ablation Study of CNN-MLP network}
Here we study the effects of changing the structure of the trainable CNN-MLP model. We fix the backbone network as the best performing model and the kernel size to $4\times4$ as obtained in Table \ref{abla_ensem}. We experimented with multiple numbers of hidden layers in the MLP each with varying number of neurons, including the case with no hidden layer at all. The results are listed in Table \ref{abla_cnnmlp}. The best result was obtained for the case when there were 2 hidden layers, with 256 and 32 neurons respectively. The input layer of the MLP depends on the output of the CNN block. From this we conclude the optimal number of filters in the CNN to be 512.

\begin{table}[h]
\centering
\begin{tabular}{l l l}
\hline
\textbf{Hidden Layers} & \textbf{Structure} & \textbf{Accuracy (\%)}\\
\hline
0 & $1024\rightarrow5$ & 57.56\\
  & $512\rightarrow5$ & 54.09\\
\hline
1 & $1024\rightarrow128\rightarrow5$ & 65.33\\
  & $512\rightarrow128\rightarrow5$ & 70.68\\
  & $256\rightarrow128\rightarrow5$ & 66.68\\
\hline
2 & $1024\rightarrow512\rightarrow128\rightarrow5$ & 82.93\\
  & $512\rightarrow256\rightarrow32\rightarrow5$ & 89.08\\
  & $256\rightarrow128\rightarrow16\rightarrow5$ & 73.10\\
  & $128\rightarrow64\rightarrow16\rightarrow5$ & 73.25\\
\hline
\end{tabular}
\caption{Ablation Study of CNN-MLP structure with ResNet 50 as the backbone and kernel size = $4\times4$.}
\label{abla_cnnmlp}
\end{table}

\subsection{Ablation study of number of feature extractors}

In this section we study the impact of the combining multiple models together. We select the three best performing models from the previous sections\footnote{We experimented with multiple combinations involving the other models as well, but the combination of the three best models produced the best results.}. We compare the performances of the models taken one at a time, two at a time, and all three at a time. Considering the practical memory constraints of most mobile devices, we limited our study to a maximum of three models taken together to restrict the total parameter count. Table \ref{abla_fe} lists the accuracies thus obtained. We can observe a clear performance improvement as we increase the number of pre-trained networks for feature extraction.

The class separability of the combinations is visualised using a TSNE \cite{van2008visualizing} plot given in Figure \ref{other}. A $t$-distributed stochastic neighbour embedding, or TSNE, is a dimensionality reduction tool that helps in visualizing the clustering ability of a model. We observe that when we take the backbones one at a time (1, 2, and 3), the models fail to form sharp clusters, thus having the lowest accuracy. The clustering capability of the model improves as we increase the number of backbone networks. Figure 4 shows the sharp clustering capability of the proposes model.

\begin{table}[h]
\centering
\begin{tabular}{l l}
\hline
\textbf{Backbone} & \textbf{Accuracy (\%)}\\
\hline
RN50 & 88.24\\
DN201 & 87.53\\
ENB5 & 82.76\\
RN50 + DN201 & 89.39\\
ENB5 + DN201 & 90.12\\
RN50 + ENB5 & 90.95\\
RN50 + DN201 + ENB5 & 92.30\\
\hline
\end{tabular}
\caption{Ablation Study of Feature Extractors for ensembling strategy with a kernel size = $4\times4$.}
\label{abla_fe}
\end{table}

\section{Comparison with the state-of-the-art}
The final results of our model is presented in this section. We provide the performance scores on the Powerline components dataset first and then compare our model with other popular models on some standard datasets.

\begin{figure}[h]
    \centering
    \includegraphics[width=0.48\textwidth]{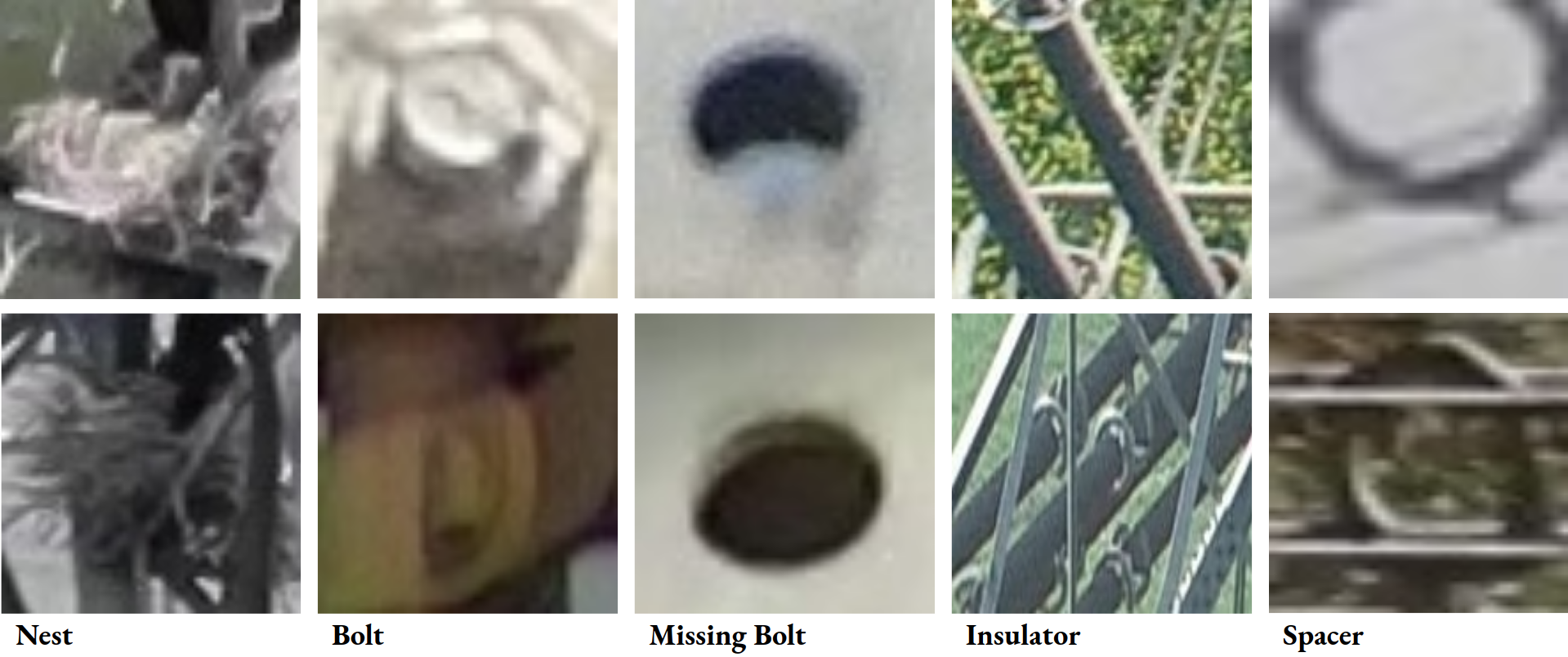} 
    \caption{Some examples of correctly classified images.}
    \label{correct}
\end{figure}

\begin{figure}[h]
    \centering
    \includegraphics[width=0.48\textwidth]{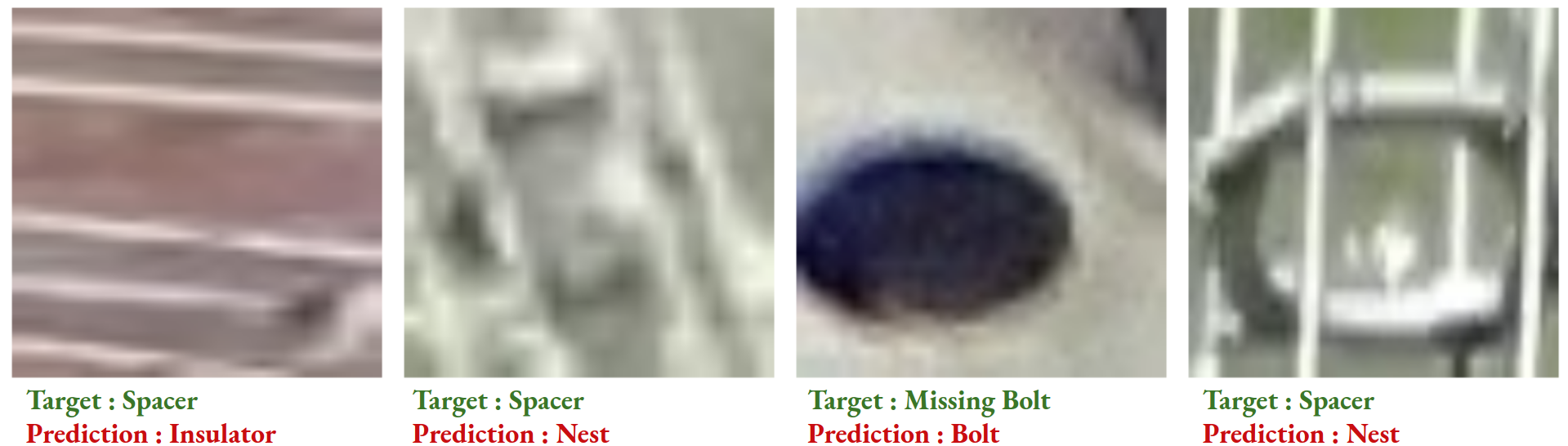} 
    \caption{Some examples of incorrectly classified images.}
    \label{incorrect}
\end{figure}

\subsection{Results on Powerline Dataset}
All the testing was done by running the model multiple times and using $k$-fold cross validation method to get an average score. Figure \ref{correct} shows some examples of images that were correctly classified by the model. It is to be noted that the images were of different resolutions. They are rescaled to the same size for display purposes.

Figure \ref{incorrect} shows some of the misclassified images. It can be observed that most of the misclassifications were due to heavy amount of noise and blur.

\begin{figure}[b]
    \centering
    \includegraphics[width=0.48\textwidth]{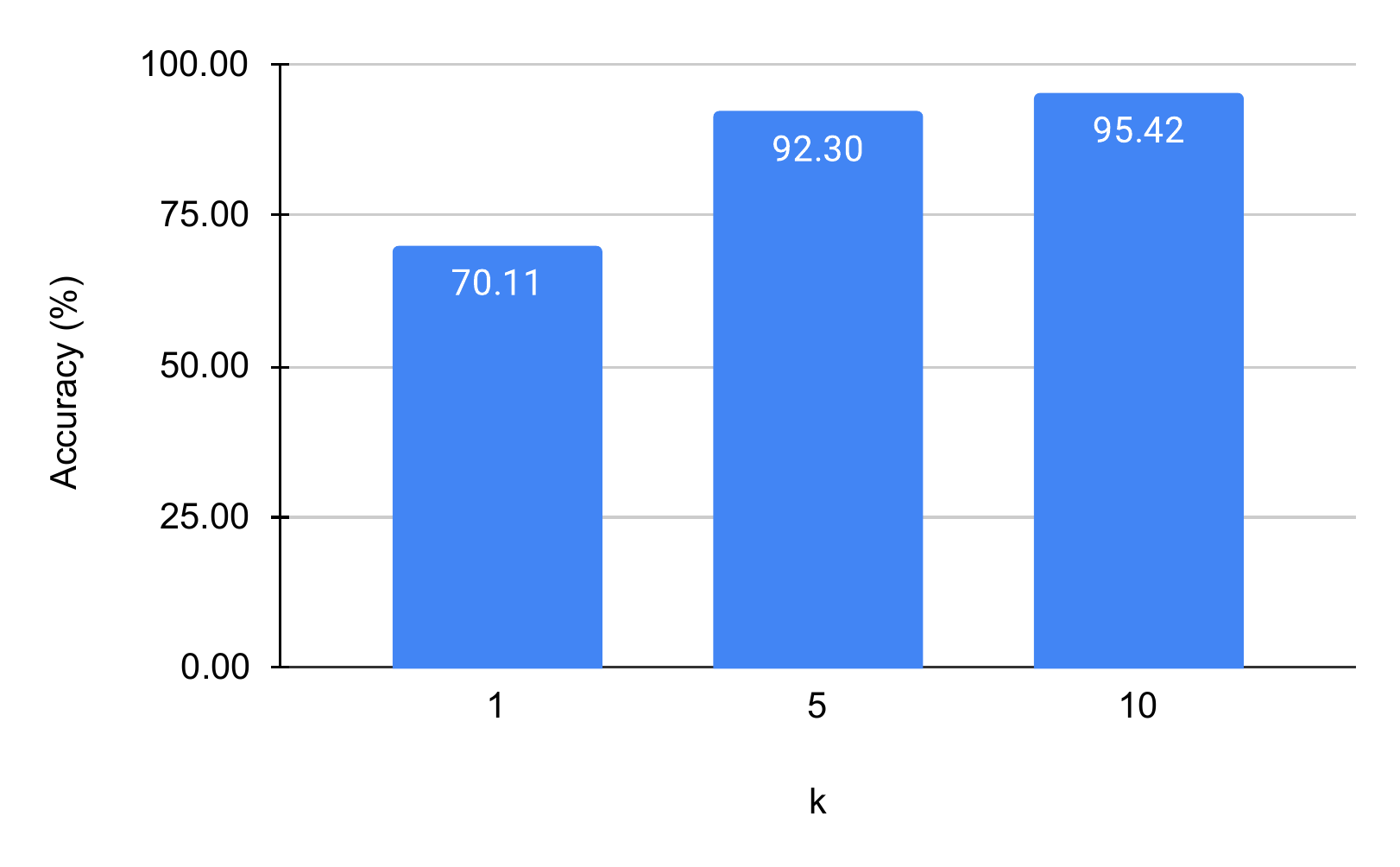} 
    \caption{Results obtained using our best model for different values of $k$ in $5$-way $k$-shot.}
    \label{k_chart}
\end{figure}
\begin{figure*}[t]
    \centering
    \includegraphics[width=1.0\textwidth]{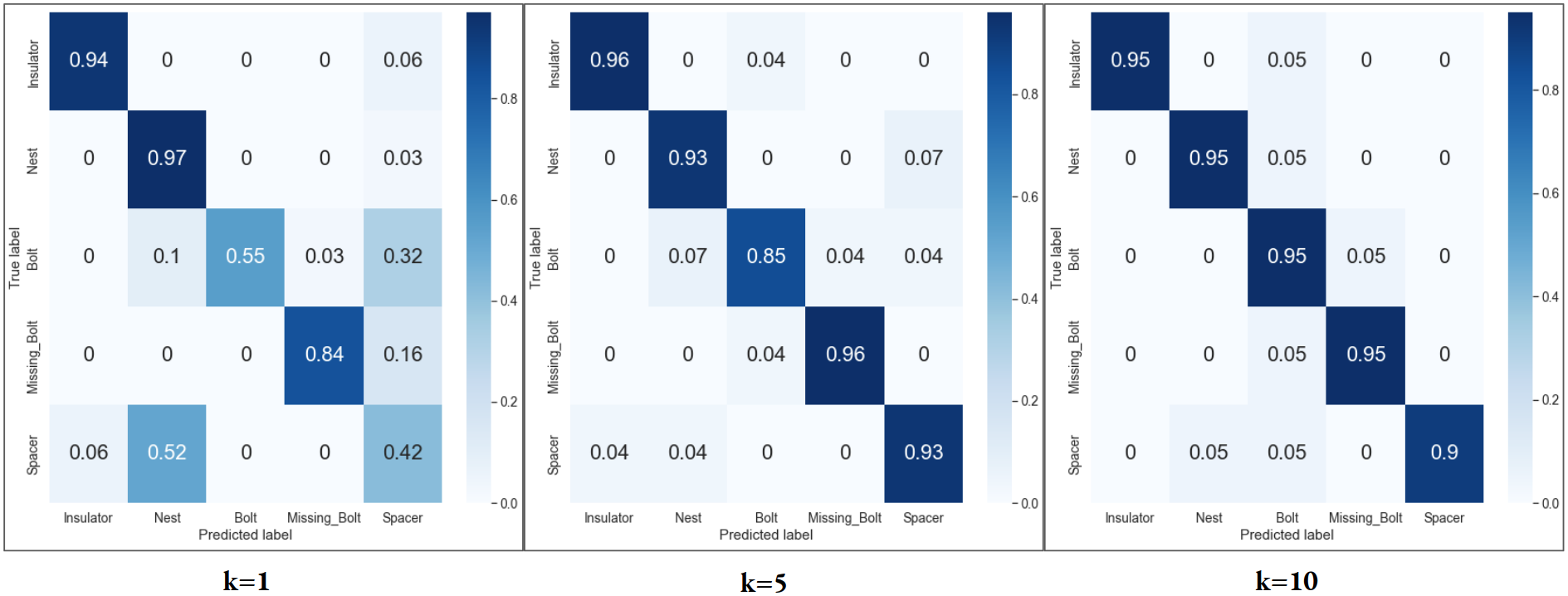} 
    \caption{Confusion matrices obtained using different values of $k$ in $5$-way $k$-shot. With increase in $k$, the matrix becomes more diagonalized implying an improve in performance.}
    \label{conf}
\end{figure*}

\begin{table*}[h]
\centering
\large
\begin{tabular}{l l l l l l l}
\hline
\textbf{Model} & \textbf{Aircraft} & \textbf{Traffic} & \textbf{Omniglot} & \textbf{Texture} & \textbf{FC100} & \textbf{VGG Flower}\\
\hline
MAML\cite{finn2017model} & 33.1 ± 0.6 & 67.4 ± 0.9 & 82.6 ± 0.7 & 56.9 ± 0.8 & 62.0 ± 0.8 & 78.0 ± 0.7\\
MatchingNet\cite{vinyals2016matching} & 33.5 ± 0.6 & 73.7 ± 0.8 & 89.7 ± 0.5 & 54.7 ± 0.7 & 59.4 ± 0.8 & 74.2 ± 0.8\\
ProtoNet\cite{snell2017prototypical} & 41.5 ± 0.7 & 75.0 ± 0.8 & 95.5 ± 0.3 & 62.9 ± 0.7 & 64.7 ± 0.8 & 86.7 ± 0.6\\
SUR\cite{dvornik2020selecting} & 45.2 ± 0.8 & 70.6 ± 0.8 & 98.7 ± 0.1 & 59.6 ± 0.7 & 67.2 ± 1.0 & 90.8 ± 0.5\\
Chowdhury et al.\cite{chowdhury2021few} & \textbf{68.9 ± 0.9} & 85.8 ± 0.7 & 98.0 ± 0.2 & 85.7 ± 0.6 & 80.5 ± 0.6 & 97.9 ± 0.2\\
\textbf{Ours}             & 65.6 ± 1.7 & \textbf{93.1 ± 0.3} & \textbf{99.0 ± 0.3} & \textbf{86.8 ± 0.6} & \textbf{91.4 ± 0.2} & \textbf{98.8 ± 0.3}\\ 
\hline
\end{tabular}
\caption{Comparative analysis of our model with the existing state-of-the-art methods for a $5$-way $5$-shot problem.}
\label{compare}
\end{table*}

Figure \ref{k_chart} lists the results obtained by our best model with the three feature extractors namely, ResNet 50, DenseNet 201, EfficientNet B5. The ensembling strategy used a kernel size of $4\times4$ and 512 filters for the CNN block, and two hidden layers ($256\rightarrow 32$) in the MLP block. The results were obtained by varying the number of training examples in each class. Figure \ref{conf} contains the confusion matrices for the three values of $k$. For $k=1$, the model was supplied with only one training image per class, explaining the sharp drop in accuracy.

\subsection{Results on Standard Datasets}

We compare our model with the existing state-of-the-art models on various standard datasets, as mentioned before, the Aircraft, Traffic, Omniglot, FC100, VGG Flower, and the Texture dataset. As we perform our hyperparameter search on the power-line anomaly dataset containing five classes only, we stick to the results for the $5$-way $5$-shot problem. Table \ref{compare} shows a detailed comparative study of our method with the existing state-of-the-art methods. For comparison we chose some of the most popular existing alternative techniques for few-shot classification, such as, MAML \cite{finn2017model}, MatchingNet \cite{vinyals2016matching}, ProtoNet \cite{snell2017prototypical}, SUR \cite{dvornik2020selecting} and the model proposed by Chowdhury et. al. \cite{chowdhury2021few}. It can be observed that under most circumstances, our method is able to outperform the model by Chowdhury et al., our inspiration, by a significant margin for most datasets.

\section{Conclusion}
In this paper we experimented a new approach for few-shot image classification. We evaluated our approach on a powerline anomaly dataset where the anomaly class was "missing bolts". We developed an ensembling technique that combines the extracted features of different pre-trained networks in a parameter efficient way. The classification accuracy obtained by training the model with a $5$-way $k$-shot support set was above 90\% for $k \geq 5$. After extensive performance evaluation with multiple combinations of feature extractors, we found that the accuracy score was obtained with a strategic combination of three specialized pre-trained networks. We visualized the class separability of our method using TSNE plots and confusion matrices and finally obtained a peak classification accuracy of 92.30\% for $5$-way $k$-shot task. The dataset used to evaluate our framework was new and challenging because it included realistic images of multiple resolutions. A major critique of our approach is the sensitivity of the accuracy on each image of the support set. For example, as the support set was randomly selected, there were cases where the all the samples under a particular class were similar to each other and failed to represent other variations thereby compromising the overall accuracy. Therefore, the selection of support set should be done with extreme care. The complete code will be made publicly available for further research.

\clearpage
\begin{acks}
All computations were performed using the resources provided by the AI Computing Facility at CSIR-CEERI, Pilani.
\end{acks}

\bibliographystyle{unsrt}
\bibliography{acmart}

\appendix

\end{document}